# VR IQA NET: DEEP VIRTUAL REALITY IMAGE QUALITY ASSESSMENT USING ADVERSARIAL LEARNING

*Heoun-taek Lim, Hak Gu Kim, and Yong Man Ro*\*

Image and Video System Lab, School of Electrical Engineering, KAIST, South Korea

**ABSTRACT**

In this paper, we propose a novel virtual reality image quality assessment (VR IQA) with adversarial learning for omnidirectional images. To take into account the characteristics of the omnidirectional image, we devise deep networks including novel quality score predictor and human perception guider. The proposed quality score predictor automatically predicts the quality score of distorted image using the latent spatial and position feature. The proposed human perception guider criticizes the predicted quality score of the predictor with the human perceptual score using adversarial learning. For evaluation, we conducted extensive subjective experiments with omnidirectional image dataset. Experimental results show that the proposed VR IQA metric outperforms the 2-D IQA and the state-of-the-arts VR IQA.

## 1. INTRODUCTION

Virtual reality (VR) and augmented reality (AR) are gaining much interest of industry, research, and customer as realistic contents in entertainment, training, education, etc. With the development of high-end head mounted display (HMD) and the production of high-quality 360 degree contents, it can allow viewers to have realistic viewing experience and interactions. While conventional 2-D images have a limited field of view (FoV), 360 degree images (i.e., *omnidirectional images* [1]) have an unlimited FoV in all direction (see Fig. 1(c)). Thus, viewers can select and see a specific portion of spherical images (i.e., *viewport*), as shown in Fig. 1(a).

Despite the advantages of VR content like an omnidirectional image, there have several challenges such as huge storage, heavy computation, and large bandwidth [2].

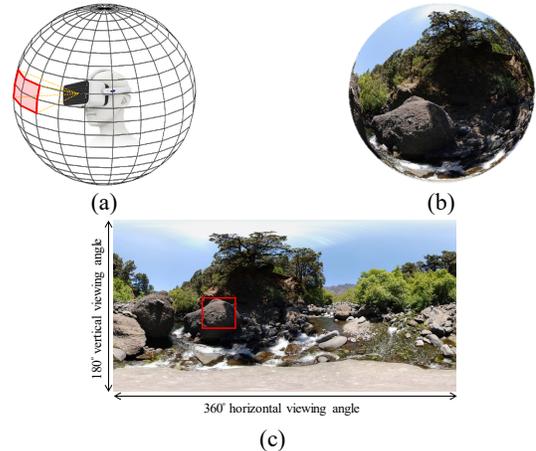

Fig. 1. (a) Viewer centered in spherical domain, (b) Spherical image, and (c) Panoramic image projected from (b) using Equirectangular projection. Red boxes in (a) and (c) indicate the viewport according to viewer's viewpoint.

In particular, delivering VR content is one of the main bottlenecks since most of omnidirectional images have a very high resolution such as 4K, 8K, and higher to cover full sphere [3], [4]. Therefore, conventional image and video coding standards for 2-D rectangle images (i.e., rectilinear image and video coding techniques) are not suitable to encode the spherical image. After mapping the spherical image (e.g., Fig. 1(b)) to a rectangular image (e.g., Fig. 1(c)), the projected rectangular image could be compressed for bitrate reduction. During this process, conventional image quality assessment (IQA) such as PSNR and SSIM [5] could be used to measure the quality of the compressed image. However, the IQA models for 2-D image (i.e., rectilinear metrics) do not consider the characteristics of projection from spherical to rectangle domain and compression distortion in panoramic images as well. So they could not elaborately measure the perceived quality of omnidirectional images. Therefore, it is necessary to devise a reliable objective metric for omnidirectional images.

Recently, there were a few works in the IQA for omnidirectional images. In [8], a spherical PSNR (S-PSNR) was proposed by computing the PSNR value between corresponding pixels in the spherical surface. Sun *et al.* proposed a weighted spherical PSNR (WS-PSNR) method considering the weights according to the pixel position in

---
\* Corresponding author (ymro@ee.kaist.ac.kr)

This work was supported by Institute for Information & communications Technology Promotion (IITP) grant funded by the Korea government(MSIT) (No.2017-0-00780, Development of VR sickness reduction technique for enhanced sensitivity broadcasting) spherical surface [9]. In [10], a PSNR-based IQA for omnidirectional image was proposed in which PSNR was computed between reference and distorted images mapped by Craster parabolic projection (i.e., CPP-PSNR). However, most of existing methods were based on 2-D IQA model for omnidirectional image. In addition, it was not validated that the models had a high correlation with human perception.

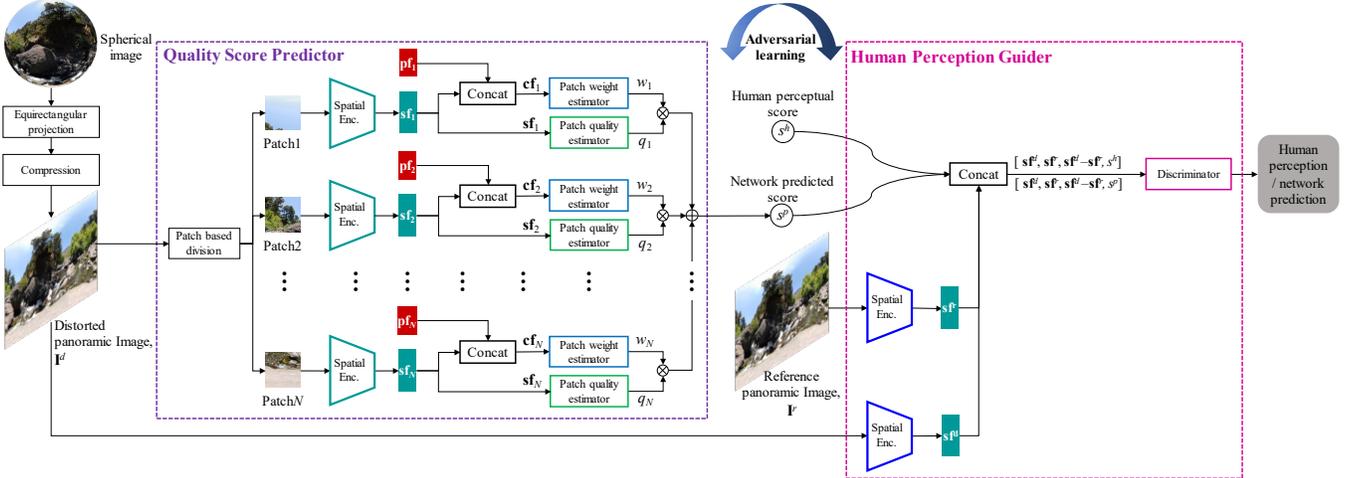

Fig. 2. Proposed VR IQA net using adversarial learning. The proposed network consists of quality score predictor and human perception guider. The quality score predictor is to measure overall visual quality scores of distorted omnidirectional images. The human perception guider is to criticize the predicted score from the quality score predictor by comparing both features of reference and distorted images.

In [1], to evaluate and compare the prediction performances of various objective metrics, subjective evaluation was conducted on omnidirectional images. The experimental results showed that current IQA models for omnidirectional images [8]-[10] could not outperform 2-D IQA models [5]-[7]. As a result, a reliable IQA model for omnidirectional images still has not been developed.

In this paper, we propose a novel deep learning-based VR IQA with adversarial learning for omnidirectional images. The main contributions of our paper are two folds:

1) We propose a new deep learning-based objective VR IQA framework using adversarial learning. In the proposed deep network, first, the proposed predictor tries to predict the quality score of omnidirectional image. The proposed critic network (called as human perception guider) discriminates the ground-truth subjective score (i.e., human perception) and the predicted score of the predictor with reference and distorted images. Then, the critic network gives a guide to the predictor based on the human rating and images. By adversarial learning between the predictor and critic, the proposed network reliably predicts the quality score.

2) For evaluation, we conducted extensive subjective assessment experiments with a large scale of omnidirectional images. We collect a total of 720 omnidirectional images containing various codec standards and compression ratio with the corresponding subjective scores in our subjective assessment.

Experimental results show that the proposed VR IQA method is highly correlated with human perception compared to the state-of-the-art VR IQA metrics.

The rest of this paper is organized as follows. In Section 2, the proposed deep learning-based VR IQA method is described. In Section 3, experiments and results are shown to demonstrate the superiority of the proposed method. Finally, conclusions are drawn in Section 4.

## 2. PROPOSED DEEP VR IQA METHOD

Fig. 1 shows the overall process of the proposed VR IQA network for omnidirectional images. The proposed network consists of the quality score predictor $P$ and human perception guider $D$. Let $\mathbf{I}^d$ and $\mathbf{I}^r$ denote the distorted and reference panoramic images, respectively. Due to the projection from the sphere to rectangle image domain, the omnidirectional image has different distortion characteristics according to spatial positions. To take into account the characteristic of omnidirectional image, the predictor $P$ encodes the spatial feature of each patch with its relative position information. Based on the features, weight and quality estimators assess weight value and quality score of each patch, respectively. Finally, the predicted quality score is obtained by combining the weight and quality score of all patches [11]. Then, the human perception guider $D$ takes the predicted score and human rating score as inputs along with distorted and reference images. It discriminates whether the input is the network predicted quality score or a human perceptual quality score. By adversarial learning [12] between $P$ and $D$, the proposed network predicts the quality score very similar to the human perceptual score.

### 2.1. VR quality score predictor

In the proposed quality score predictor, the distorted image, $\mathbf{I}^d$, divided into the 256x256 $N$ patches. Let $\mathbf{sf}_i$, $\mathbf{pf}_i$ and $\mathbf{cf}_i$ denote the spatial, position and combined features of the $i$-th image patch, respectively. $\mathbf{sf}_i$ is encoded from a spatial encoder which is the pre-trained ResNet-50 [13]. $\mathbf{sf}_i \in \mathbf{R}^{2048}$ is the feature vector of the global average pooling layer in ResNet-50. $\mathbf{pf}_i = [x_i-x_d, y_i-y_d]^T \in \mathbf{R}^2$ is relative position information between the center points of both the $i$-th image patch and the entire image. $\mathbf{cf}_i \in \mathbf{R}^{2050}$ is defined as the concatenation of $\mathbf{sf}_i$ and $\mathbf{pf}_i$.

The weight of the $i$-th image patch, $w_i$, is derived from $\mathbf{cf}_i$ by the proposed weight estimator. The proposed weight estimator reliably produces $w_i$ by considering the visual and spatial position information, $\mathbf{cf}_i$. The image patch quality, $q_i$, is derived from $\mathbf{sf}_i$ through the patch quality estimator. The patch quality estimator evaluates the objective quality of each patch by considering the visual characteristic, $\mathbf{sf}_i$. The weight and quality estimators consist of 4 FC (fully connected) layers, which are FC-512, FC-64, FC-8, and FC-1. Finally, the predicted quality score for distorted image, $\hat{s}^h$ can be written as

$$P(\mathbf{I}^d) = \hat{s}^h = \frac{\sum_i^N w_i q_i}{\sum_i^N w_i}. \quad (1)$$

## 2.2. Human perception guider

To enhance the prediction performance of the proposed VR quality score predictor during training, the human perception guider is proposed. It takes the predicted quality score by our predictor, $\hat{s}^h$ and ground-truth human perceptual quality score, $s^h$ as inputs (we call them $s$) along with distorted and reference images, $\mathbf{I}^d$ and $\mathbf{I}^r$. $\mathbf{sf}^d$ and $\mathbf{sf}^r$ denote the spatial features of $\mathbf{I}^d$ and $\mathbf{I}^r$, respectively. They are the feature vectors of the global average pooling layer in ResNet-50. After that, $\mathbf{sf}^d$, $\mathbf{sf}^r$, $\mathbf{sf}^d - \mathbf{sf}^r$, and the quality score, $s$ are concatenated as an input of discriminator. The discriminator is composed of 4 FC layers, which are FC-512, FC-64, FC-8, and FC-1. Its output represents the conditional probability of the quality score, given $\mathbf{I}^d$ and $\mathbf{I}^r$, which can be written as

$$D(s | \mathbf{I}^d, \mathbf{I}^r) = p(s \text{ is human perceptual score} | \mathbf{I}^d, \mathbf{I}^r). \quad (2)$$

## 2.3. Training the proposed VR IQA network with adversarial learning

Proposed quality score predictor $P$ and human perception guider $D$ are trained with adversarial learning [12]. During the training, $D$ is trained to discriminate the human perceptual score, $s^h$ and the network predicted score, $\hat{s}^h$ by comparing $\mathbf{I}^d$ and $\mathbf{I}^r$. On the other hand, $P$ is trained to produce, $\hat{s}^h$ which is similar to $s^h$. As a result, $D$ cannot distinguish $\hat{s}^h$ from $s^h$. Through this training strategy, $P$ with only $\mathbf{I}^d$ has the effect of the FR-IQA by $D$ with $\mathbf{I}^r$.

To that end, we design a novel objective function for adversarial learning in order to find the optimal $P$ and $D$. It can be written as

$$\min_P \max_D \{V(P,D) := (\hat{s}^h - s^h)^2 - \lambda(J(D(s^h | \mathbf{I}^d, \mathbf{I}^r),1) + J(D(\hat{s}^h | \mathbf{I}^d, \mathbf{I}^r),0))\}, \quad (3)$$

where $J(p,q) = -q \ln p - (1-q) \ln(1-p)$ is a binary cross-entropy loss. $\hat{s}^h$ is equal to $P(\mathbf{I}^d)$. The Eq. (3) can be decomposed into two optimization problems.

$$\min_P \{L_P := (P(\mathbf{I}^d) - s^h)^2 + \lambda J(D(P(\mathbf{I}^d) | \mathbf{I}^d, \mathbf{I}^r),1)\}, \quad (4)$$

$$\min_D \{L_D := J(D(s^h | \mathbf{I}^d, \mathbf{I}^r),1) + J(D(P(\mathbf{I}^d) | \mathbf{I}^d, \mathbf{I}^r),0)\}. \quad (5)$$

In the two-player minimax game with $V(P, D)$, the $D$ is trained to maximize the probability of determining the human perception class from both of ground-truth human rating and predicted score of $P$. Simultaneously, the $P$ is trained to minimize the difference between human rating score and predicted score, and maximize the probability that $D$ assigns the network prediction class to the predicted score by minimizing the cross-entropy loss.

## 3. EXPERIMENTS AND RESULTS

### 3.1. Dataset generation

For training of our model and performance evaluation, we utilized SUN360 database [14], which is a large scale of 360 degree images represented in equirectangular projection with 9104 x 4552 pixels. We randomly selected a total of 60 omnidirectional images from indoor (30 images) and outdoor (30 images) scenes. In our experiment, in order to match the resolution of our HMD (described in Section 3.3), the original high resolution omnidirectional images were down-sampled using bi-cubic interpolation to a 2048 x 1024 pixels, which were used as a reference. Then, to generate the distorted images, we compressed the 60 omnidirectional images with 2048 x 1024 pixels using three widely used codec standards, which are JPEG [15], JPEG 2000 [16], and HEVC [17]. In our experiment, we selected four different bit rates, which were 0.5, 1.0, 1.5, and 2.0 bits per pixel [bpp] for each compression. We utilized FFmpeg [18] library to compress the omnidirectional images. All the compressed images were decompressed to produce the distorted omnidirectional images. As a result, a total number of 720 omnidirectional images (720 images = 60 scenes x 3 codec standards x 4 bit rates) were obtained for evaluation.

### 3.2. Experimental setup

In this paper, our experiments were performed on a PC with Intel Xeon CPU E5-1660 v4 @ 3.20GHz, 32 GB RAM, NVIDIA GTX 1080Ti, and TensorFlow [19]. We used the $N$=32 patches with a size of 256x256 pixels. Batch size=6, learning rate=2e-4 and $\lambda$=100 were adapted for training. For evaluation, we used 5-fold cross validation.

Table 1. Prediction performances comparison

| Objective metrics | | PLCC | SROCC | RMSE |
|---|---|---|---|---|
| PSNR | | 0.6983 | 0.6794 | 12.8791 |
| SSIM [5] | | 0.7301 | 0.7259 | 12.2954 |
| MS-SSIM [6] | | 0.7383 | 0.7516 | 12.1364 |
| VIFp [7] | | 0.8124 | 0.7907 | 10.4909 |
| S-PSNR [8] | | 0.5316 | 0.5470 | 15.2403 |
| WS-PSNR [9] | | 0.5270 | 0.5172 | 15.2917 |
| CPP-PSNR [10] | | 0.5185 | 0.5347 | 15.3850 |
| **Proposed VR IQA NET** | without Critic | 0.8516 | 0.8227 | 9.4313 |
| | with Critic | **0.8721** | **0.8522** | **8.8048** |

### 3.3. Subjective assessment experiment

A total of 15 subjects participated in our subjective VR experiment. In our experiment, Oculus Rift CV1 with the *Oculus 360 Photos* was used to display the omnidirectional images. All subjects were seated on a rotatable chair. All experimental settings followed the guideline, ITU-R BT.500-13 [20] and BT.2021 [21].

In our subjective assessment experiment, we measured the overall visual quality using the single stimulus continuous quality evaluation (SSCQE) [20]. Each stimulus was displayed for 15 s through the Oculus Rift CV1. Then, a mid-gray image was presented for 5 s for resting. The subjects scored their perceived quality in the continuous scale range of 0-100, divided into five grades: excellent, good, fair, poor, and bad. Our subjective assessment consisted of 6 sessions. Each session was conducted on a different day. During each session, the subjects could immediately stop and take a rest if they felt difficult to continue due to excessive visual fatigue.

### 3.4. Prediction performance

To evaluate the prediction performance of the proposed objective VR IQA, three performance measures were employed: Pearson linear correlation coefficient (PLCC), Spearman rank order correlation coefficient (SROCC), and root mean square error (RMSE) between subjective MOS values obtained by our subjective assessment and objective MOS values obtained by objective quality assessments.

For performance comparison, seven existing objective metrics were used. Four metrics were 2-D IQA models, which are PSNR, SSIM [5], MS-SSIM [6], and VIFp [7]. Three metrics were state of the art of VR IQA, which are S-PSNR [8], WS-PSNR [9], and CPP-PSNR [10].

Table 1 shows the prediction performance of the proposed method and existing objective metrics. In Table 1, the proposed method *without Critic* indicates that we only learned quality score predictor except for the human perception guider. As shown in Table 1, the proposed VR IQA metric achieved the highest correlation and the lowest RMSE with subjective MOS, compared to those of existing 2-D metrics and the state-of-the art VR IQA metrics.

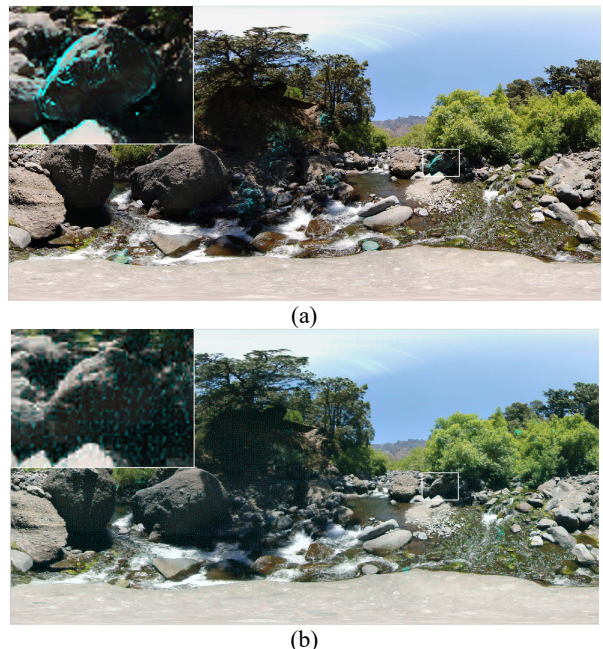

(a)

(b)

Fig. 3. Visualization of the proposed VR IQA Net. The distorted images and the gradients of the predicted score (cyan) with (a) the highest MOS: 70.85 (HEVC with 2.0 bpp) and (b) the lowest MOS: 7.14 (JPEG with 0.5 bpp). Note that top-left shows the magnified parts of the local regions (white boxes).

### 3.5. Interpretation of the proposed deep VR IQA model

Fig. 3 visualizes the gradients of activation that most affect the predicted score by the proposed network using the guided back-propagation [22]. Fig. 3(a) shows that some specific objects near the center had a significantly high activation on the score of the network in the high quality image. On the other hands, Fig. 3(b) shows that, in the low quality image, overall gradients of the proposed network were mainly activated along the blocky artifacts. The experimental results indicated that the proposed network focused on the compression artifacts for low quality images while focused on the saliency objects for high quality images.

## 4. CONCLUSIONS

In this paper, we proposed a novel VR-IQA NET using adversarial learning for automatically assessing the image quality of VR content. The proposed quality score predictor could reliably predict the quality score of the omnidirectional images by considering the spatial characteristics of the projection from sphere to rectangle domain. The proposed human perception guider could make the predictor to more correct by comparing the predicted score of the predictor and the human perceptual score with adversarial learning. Experimental results showed that the proposed VR IQA method was strongly correlated with human perception. In particular, we interpreted how the proposed network predicted the quality score by visualizing the activated regions by our network.